\documentclass[10pt,twocolumn,letterpaper]{article}

\usepackage{iccv}
\usepackage{times}
\usepackage{epsfig}
\usepackage{graphicx}
\usepackage{amsmath}
\usepackage{amssymb}

\usepackage{tabularx} 
\newcolumntype{C}{>{\centering\arraybackslash}X}
\usepackage{booktabs}

\makeatletter\let\captiontemp\@makecaption\makeatother
\usepackage[font=small]{subcaption}
\makeatletter\let\@makecaption\captiontemp\makeatother

\usepackage{sg-macros}

\usepackage[pagebackref=true,breaklinks=true,letterpaper=true,colorlinks,bookmarks=false]{hyperref}
\usepackage[accsupp]{axessibility}  

\iccvfinalcopy 

\def\iccvPaperID{3553} 

\ificcvfinal\fi

\begin{document}

\title{Spatially Conditioned Graphs for Detecting Human--Object Interactions}

\author{Frederic Z. Zhang$^{1, 3}$ \quad Dylan Campbell$^{2, 3}$ \quad Stephen Gould$^{1, 3}$ \\
$^1$The Australian National University \quad $^2$University of Oxford \\ $^3$Australian Centre for Robotic Vision\\
{\tt\small \{firstname.lastname\}@anu.edu.au \quad dylan@robots.ox.ac.uk} \\
{\tt\small \href{https://github.com/fredzzhang/spatially-conditioned-graphs}{https://github.com/fredzzhang/spatially-conditioned-graphs}}
}
\maketitle
\ificcvfinal\thispagestyle{empty}\fi

\begin{abstract}
   We address the problem of detecting human--object interactions in images using graphical neural networks. Unlike conventional methods, where nodes send scaled but otherwise identical messages to each of their neighbours, we propose to condition messages between pairs of nodes on their spatial relationships, resulting in different messages going to neighbours of the same node. To this end, we explore various ways of applying spatial conditioning under a multi-branch structure. Through extensive experimentation we demonstrate the advantages of spatial conditioning for the computation of the adjacency structure, messages and the refined graph features. In particular, we empirically show that as the quality of the bounding boxes increases, their coarse appearance features contribute relatively less to the disambiguation of interactions compared to the spatial information. Our method achieves an mAP of 31.33\% on HICO-DET and 54.2\% on V-COCO, significantly outperforming state-of-the-art on fine-tuned detections.
\end{abstract}

\section{Introduction}

\begin{figure}[t]\centering
	\begin{subfigure}[t]{\linewidth}
	   \centering
		\includegraphics[width=0.8\linewidth]{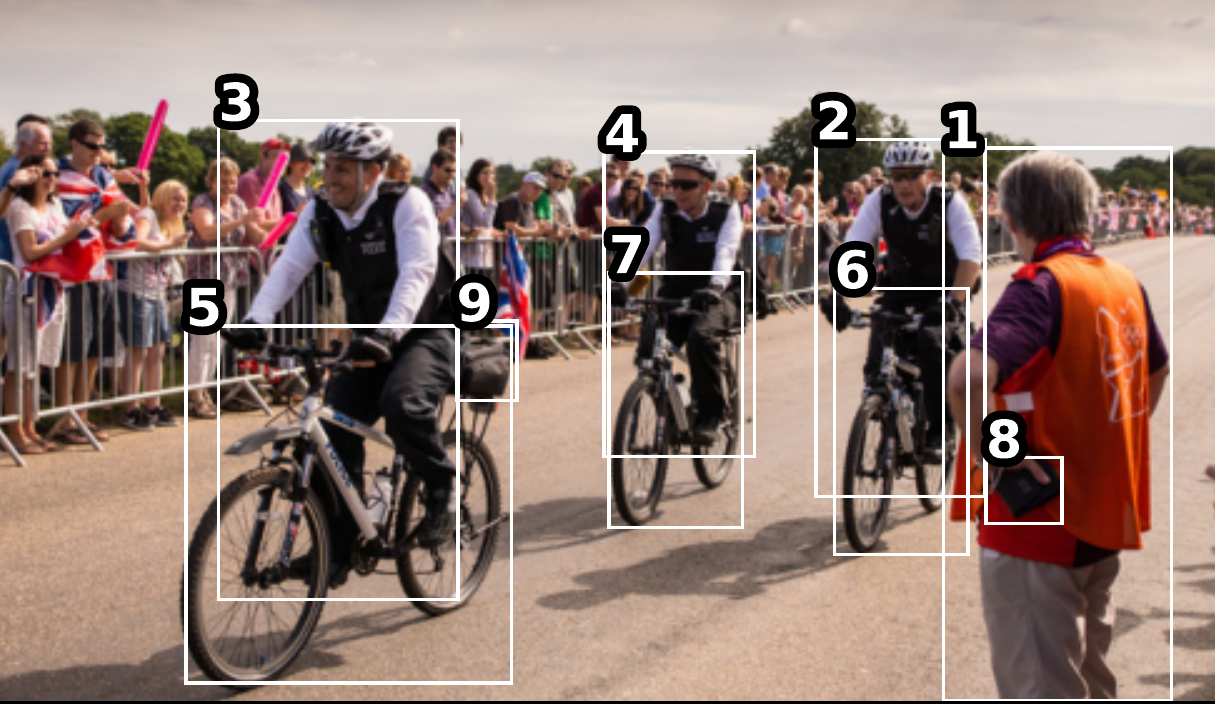}
	   \caption{An image with detected human and object instances}
	   \label{fig:riding_bikes}
	\end{subfigure}
	
	\begin{subfigure}[t]{\linewidth}
	   \centering
		\includegraphics[width=0.49\linewidth]{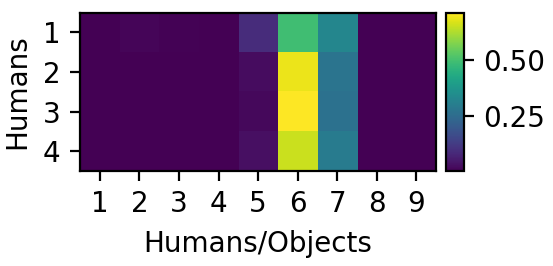}
	   \includegraphics[width=0.49\linewidth]{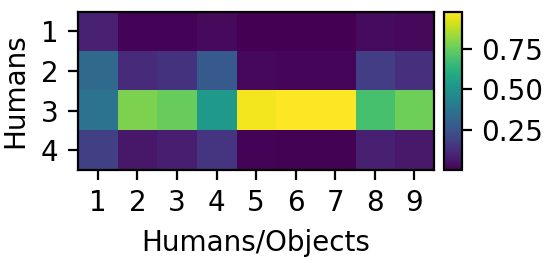}
	   \caption{Adjacency matrices computed with appearance features, normalised by rows (left) and columns (right)}
	   \label{fig:adj_baseline}
	\end{subfigure}
	
	\begin{subfigure}[t]{\linewidth}
	 \centering
	  \includegraphics[width=0.49\linewidth]{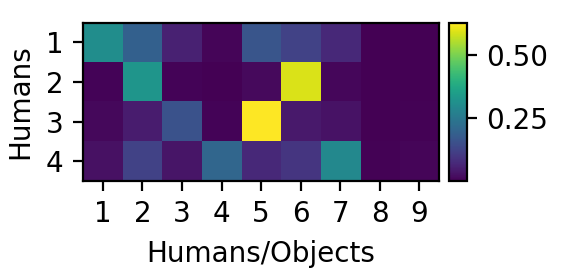}
	 \includegraphics[width=0.49\linewidth]{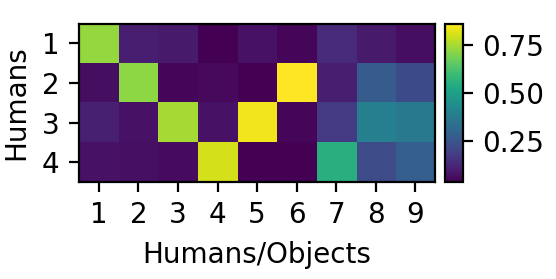}
	 \caption{Adjacency matrices computed with spatial conditioning, normalised by rows (left) and columns (right)}
	 \label{fig:adj}
	  \end{subfigure}
	\newline
	\caption{Many images contain far more non-interactive human--object pairs than interactive ones (\subref{fig:riding_bikes}). Correct inference of the interaction type and the correspondences requires a combination of appearance and spatial information. When using appearance features only, the adjacency matrix for a graphical neural network tends to be dominated by a few salient objects (\subref{fig:adj_baseline}). Since messages from each node to its neighbours are identical apart from an adjacency scaling, this leads to the node features being dominated by those of the most salient objects, confusing the classifier. With spatial conditioning, the adjacency matrix is able to reflect the inherent interactive pairs without explicit supervision (\subref{fig:adj}).}
	\label{fig:teaser}
 \end{figure}

The task of detecting human--object interactions (HOIs) requires localising and describing pairs of interacting humans and objects. In particular, an HOI is defined as a (subject, predicate, object) triplet, following the definition of visual relations from Lu \etal~\cite{lu2016}, where the subject and object are typically represented as labelled bounding boxes. For HOI triplets, the subject is always a human, so the interactions of interest simplify to pairs of predicates and objects, \eg, \textit{riding a horse} or \textit{sitting on a bench}.

Since the output representations are inherently similar, HOI detection is most often approached as a downstream task of object detection. Given a set of object detections from an image, one may construct candidate human--object pairs by exhaustively matching between the detected human and object instances. Indeed, the vast majority of previous works~\cite{chao2018, gao2018, gupta2019, li2019, qi2018, peyre2019, ulutan2020, gao2020, wang2020} use an off-the-shelf object detector~\cite{ren2015} as a preprocessing stage. We take the same approach, leveraging the success of modern object detectors. While this converts the HOI detection task into the simpler HOI recognition task on a set of candidate human--object pairs, it is still far from being solved.

Recognising HOIs is extremely challenging. While image recognition discriminates between scene types~\cite{zhou2017} or prominent object types~\cite{russ2015}, focusing on the holistic understanding of an image, HOI recognition requires an understanding of the interactions between specific humans and objects at a much finer level. This requires reasoning about the subtle relationships between the instances as well as their contexts. This is particularly necessary when there are multiple human--object pairs with the same interaction type, where the model needs to correctly infer the interaction type and the correspondences between the individual instances.
In addition, many interactions do not have strong visual cues and can be quite abstract, such as \textit{buying an apple} or \textit{inspecting a boat}. This poses a big challenge for standard CNNs, which excel at recognising physical qualities such as texture and shape. HOI detection demands a more sophisticated architecture capable of performing logical reasoning, not merely recognising the visual cues of the humans and objects of interest. The complexity and ambiguity of the problem is such that even humans can fail to correctly recognise HOIs in images, despite our ability to reason about visual cues and spatial relationships. Following prior work, we make use of graphical models to model these interrelationships and perform structured prediction.

\begin{table}[t]\small
	\caption{The use of appearance (A) and spatial (S) modalities at different stages of the graphical model, in recent HOI works. Refinement refers to late-stage fusion that takes place after message passing and fuses the graph features with other modalities.}
	\label{tab:modalities}
	\setlength{\tabcolsep}{1pt} 
	\begin{tabularx}{\linewidth}{l C C C}
		\toprule
		& \textbf{Adjacency} & \textbf{Message} & \textbf{Refinement}\\
		\textbf{Methods} & \textbf{(early fusion)} & \textbf{(mid fusion)} & \textbf{(late fusion)}\\
		\midrule
		GPNN~\cite{qi2018} & A & A & -- \\
		Wang \etal~\cite{wang2020} & A, S & A & A, S \\
		DRG~\cite{gao2020} & S & S & -- \\
		VSGNet~\cite{ulutan2020} & A & A & A, S \\
		Ours & A, S & A, S & A, S \\
		\bottomrule
	\end{tabularx}
\end{table}

Since humans and objects in an image play different roles in the interactions, we build a bipartite graph to characterise these interrelationships, wherein each human node is connected to each object node. As is intuitive, we use the appearance features of a detected instance as the node encoding, be it a person or an object. Edge encodings, however, have been under-explored in the HOI detection problem. Previous works~\cite{qi2018,wang2020} take the appearance feature extracted from the minimum covering rectangle of the human and object boxes as the edge encoding. This representation does not necessarily encode the spatial relationships between a human--object pair, and there could be additional objects in the tight box other than the intended pair. Instead, we use explicitly learned spatial representations as the edge encodings. To shed some light on their significance, let us consider the example shown in Figure~\ref{fig:riding_bikes}. Graphical models allow the propagation of contextual information between nodes. In this instance, each human node will receive information suggesting the presence of bikes. However, conventional algorithms send identical messages from a node to its neighbours, with the sole variable being a learnable weight that characterises the connectivity. And Figure~\ref{fig:adj_baseline} shows that this connectivity matrix fails to identify correct human--bike pairs with only appearance information, which causes confusion when distinguishing between all putative human--bike pairs. As such, we contend that it is crucial to incorporate spatial information to regulate the message passing procedure.
Our intuition is that, with spatial conditioning, each human node receives information of the presence of a bike \textit{and} its relative location. Therefore, the interaction \textit{riding a bike} could potentially be suppressed for a human instance if all bikes in the image are, say, to its left, as opposed to being directly under it.

Our primary contribution is a spatially conditioned message passing algorithm that renders outgoing messages which are dependent on the receiving nodes. For our bipartite graph, the algorithm also passes anisotropic messages across the bipartition. Furthermore, we extend the spatial conditioning mechanism to other parts of the graph---the computation of the adjacency structure and the refinement of the graph features---through a proposed multi-branch fusion module. While previous works have also combined appearance and spatial modalities at these two stages of the network as shown in Table~\ref{tab:modalities}, our approach is consistent at each fusion stage and, in particular, gains significant performance improvements from using both modalities during message passing.
Our secondary contribution is an analysis of the relative significance of the different modalities. We empirically show that as detection quality improves, the importance of the coarse appearance features decreases compared to that of the spatial information. We obtain state-of-the-art performance on the HICO-DET~\cite{chao2018} and V-COCO~\cite{gupta2015} datasets, establishing a new benchmark for detecting human--object interactions.

\section{Related Work}

\begin{figure*}[t]\centering
	\includegraphics[height=0.28\linewidth]{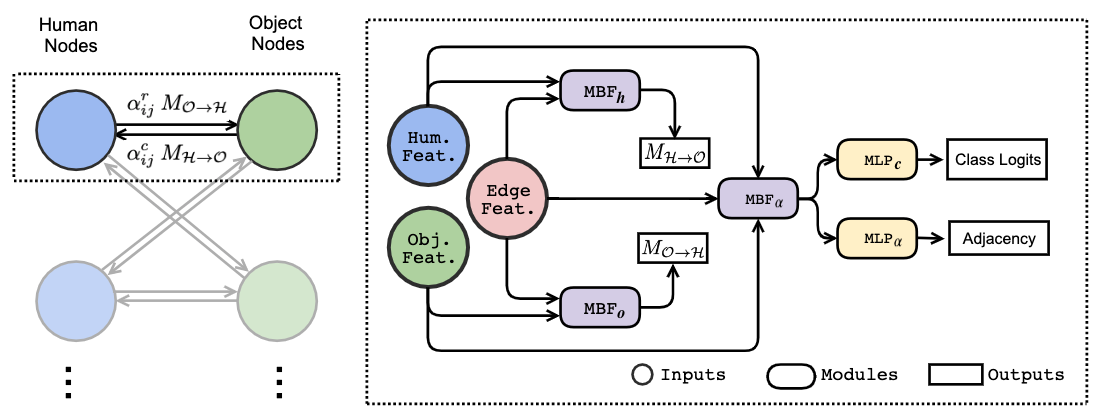}
    \caption{Diagram of proposed bipartite graph structure (left) and message passing algorithm (right). The graph structure and its connectivity is shown on the left, specifically highlighting the directed edges and anisotropic message passing. On the right, we zoom in on a particular pair of nodes and illustrate the computation of adjacency (Eq.~\ref{eq:adj}), messages (Eq.~\ref{eq:msg_h}, \ref{eq:msg_o}) and class logits. For better clarity, we intentionally leave the update functions out of the diagram and refer the readers to the equations (Eq.~\ref{eq:update_h}, \ref{eq:update_o}).
    }
   \label{fig:pipeline}
\end{figure*}

\textbf{The HOI detection pipeline} has significant overlap with that of object detection. Analogous to two-stage object detectors, a common approach is to first generate human--object pair proposals and then classify their interactions. Specifically, Faster R-CNN~\cite{ren2015} has been used in many preceding works~\cite{chao2018, gao2018, gupta2019, li2019, qi2018, peyre2019, ulutan2020, gao2020, wang2020,hou2020,li2020} to generate objects, each of which is associated with a predicted class and a confidence score. Afterwards, with appropriate filtering, human--object pairs are constructed exhaustively from the remaining detections. That is, each human instance will be paired up with each object instance. The rest of the pipeline varies, but typically employs a network with multiple streams to exploit different modalities of information. For instance, Chao \etal~\cite{chao2018} proposed a three-branch architecture to process the human and object appearance features and their pairwise spatial relationships. Different to many previous works, Liao \etal~\cite{liao2020} presented a proposal-free HOI detection pipeline, where interactions are directly detected as keypoints. Such a keypoint represents the centre of the minimum covering rectangle for a human--object pair engaged in the predicted interaction. Positions of the human and object instances are obtained by regressing the displacements with respect to the detected interaction keypoint, similar to CornerNet~\cite{law2018}, a keypoint-based object detector. Instead, we adopt the ubiquitous approach of using an off-the-shelf detector, due to their high performance and stability, and focus on improving the classification performance given a set of detections.

\textbf{The choice of features} has undergone significant development in recent research on HOI detection. Chao \etal~\cite{chao2018} used RoIPool~\cite{girshick2015} to extract human and object appearance features and handcrafted a two-channel binary mask to encode the pairwise spatial relationships. While RoIAlign~\cite{he2017} is now used in preference to RoIPool for appearance feature extraction, the binary mask is still widely used~\cite{gao2018,gao2020,li2019,ulutan2020,wang2020}. However, Gupta \etal~\cite{gupta2019} argued that a handcrafted spatial feature is a more effective way to encode the spatial relationships, explicitly exposing the coordinates of the bounding box pairs, the intersection over union, the aspect ratios, \etc. They and others~\cite{li2019,zhou2019,wan2019} also proposed the use of human pose as additional information, which leads to some success in a few previous methods. We observe similar benefits to using handcrafted spatial encodings, but do not make use of human pose information in this work. Instead, we focus on showing how structured architectures can best exploit appearance and spatial information to disambiguate human--object interactions.

\textbf{Graphical models} were introduced to HOI detection by Qi \etal~\cite{qi2018}. They proposed a fully-connected graph with detected human and object instances as nodes. The node features are initialised with box appearance features and iteratively updated with a message passing algorithm. Wang \etal~\cite{wang2020} argued that the graph should take into consideration the fact that there are two sets of heterogeneous nodes, that is, the human nodes and object nodes. Thus, message passing between homogeneous nodes (intra-class messages) should be modelled differently from that between heterogeneous nodes (inter-class messages). Gao \etal~\cite{gao2020} also took advantage of the heterogeneity in nodes by constructing separate human-centric and object-centric graphs. They modelled human--object pairs as nodes, and employed the pairwise spatial relationships as node encodings. Last, Ulutan \etal~\cite{ulutan2020} proposed a bipartite graph in addition to a visual branch, which makes use of the appearance features of human--object pairs and the global scene. Most of the previous methods use both appearance and spatial modalities in graphical models as shown in Table~\ref{tab:modalities}. However, the messages in all of their graphical models contain only one of the two modalities. Furthermore, the messages sent from a node to its neighbours are identical except weighted by adjacency values, which is what makes this work distinct.


\section{Spatially Conditioned Graphs}

To reason jointly about the appearance and spatial information of an image, we propose a graph neural network for detecting human--object interactions.
The structure of the graph is shown in Figure~\ref{fig:pipeline}. To obtain an initial set of detections $\{d_{i}\}_{i=1}^{n}$ for each image, we run an off-the-shelf object detector and apply appropriate filtering. We use Faster R-CNN~\cite{ren2015}, although our model is detector agnostic. The detections are given by the tuple $d_{i}=(\bb_{i}, s_{i}, c_{i})$, with bounding box coordinates $\bb_{i} \in \reals^{4}$, confidence score $s_{i} \in [0,1]$ and predicted object class $c_{i} \in \cK$, where $\cK$ is the set of object categories dependent on the dataset.

\subsection{A Bipartite Graph Structure}

We denote the bipartite graph as $\cG=(\cH, \cO, \cE)$, where $\cH=\{d_{i} \mid c_{i} = \text{``person"}\}$, $\cO=\{d_{i} \mid c_{i} \neq \text{``person"}\}$, and $\cE$ is the set of edges, such that all vertices on one side of the bipartition are densely connected to those on the other. The node encodings are initialised with appearance features extracted using RoIAlign~\cite{he2017}, and the edge encodings are computed as handcrafted feature vectors. We start by encoding the rudimentary spatial information: centre coordinates of the bounding boxes, widths, heights, aspect ratios and areas, all normalised by the corresponding dimension of the image. To characterise the pairwise relationships, we also include the intersection over union, the area of the human box normalised by that of the object box, and a directional encoding given by $[\scalebox{0.9}{\texttt{ReLU}}(d_{x}) \;\;\; \scalebox{0.9}{\texttt{ReLU}}(-d_{x}) \;\;\; \scalebox{0.9}{\texttt{ReLU}}(d_{y}) \;\;\; \scalebox{0.9}{\texttt{ReLU}}(-d_{y}) ]$,
where $d_{x}$ and $d_{y}$ are the differences between centre coordinates of the human and object boxes normalised by the dimensions of the human box. This gives us the pairwise spatial encoding $\bp \in \reals_{+}^{18}$. Following the practice of Gupta \etal~\cite{gupta2019}, we concatenate the spatial encoding with its logarithm, allowing the network to learn second and higher order combinations of different terms. For numerical stability, a small constant $\epsilon > 0$ is added before taking the logarithm, which gives $\bp \oplus \log(\bp + \epsilon)$ as the pairwise spatial features.

To initialise the human and object nodes, the respective appearance features are mapped to a lower dimension with a multilayer perceptron (MLP) to get the node encodings $\bx_{i}^{0}, \by_{j}^{0} \in \reals^n$ for indices $i \in \{1,..., \vert \cH \vert\}$, $j \in \{1,..., \vert \cO \vert\}$ and time step $t=0$.
Similarly, the edge encoding $\bz_{ij} \in \reals^n$ is obtained by mapping the pairwise spatial features to the same dimension using another MLP. The edge encodings are constant during message passing. We define our bi-directional message passing updates as
\begin{align}
   \bx_{i}^{t+1} &= \text{LN}\!\left(\bx_{i}^{t} + \sigma\!\left( \sum_{j=1}^{\vert \cO \vert} \alpha_{ij}^{r} \: M_{\cO \rightarrow \cH}(\by_{j}^{t}, \bz_{ij})\right)\right) \label{eq:update_h}
   \\
   \by_{j}^{t+1} &= \text{LN}\!\left(\by_{j}^{t} + \sigma\!\left( \sum_{i=1}^{\vert \cH \vert} \alpha_{ij}^{c} \: M_{\cH \rightarrow \cO}(\bx_{i}^{t}, \bz_{ij})\right)\right) \label{eq:update_o},
\end{align}
where LN denotes the LayerNorm operation~\cite{ba2016}, $\sigma$ is the activation function (ReLU) and $\alpha$ is an adjacency weight between nodes. Notably, the message function $M$ is, by design, anisotropic in that it has different parametrisations for different directions. This design allows nodes to send different message tailored to the type of receiving nodes.

\subsection{Spatial Conditioning}

Appearance and spatial features constitute the two most important sources of information in the disambiguation of complex interactions. However, in all previous works~\cite{qi2018,wang2020,gao2020,ulutan2020}, messages between nodes contain only one of the two modalities, and each node sends identical messages to its neighbours, modulo an adjacency scaling. We believe that this limits the representation power of the graphical model significantly. To this end, we propose to condition the messages between nodes on their spatial relationships, which allows messages to express the relative location of the human or object, not just their presence. To do so, we fuse the edge encoding and the node encoding (of the sender) by taking the elementwise product. We justify this design choice in the ablation analysis in Section~\ref{sec:ablation}.

We extend this strategy to two other parts of the graph. First, we apply spatial conditioning to compute the adjacency matrix. This allows the learned graph connectivity to also take into account the spatial relationship. As a result, it is able to infer the interactive human--object pairs as shown in Figure~\ref{fig:adj}. Second, we apply spatial conditioning to obtain the representations for human--object pairs. That is, after message passing is finished, we concatenate the graph features of each human--object pair, conditioned on their edge encoding.
Our model therefore consistently applies spatial conditioning to compute the adjacency matrix, the messages, and the final pairwise features, which corresponds to early, mid and late fusion between the modalities.

\subsection{Multi-Branch Fusion}

\begin{figure}[t]\centering
	\includegraphics[width=0.7\linewidth]{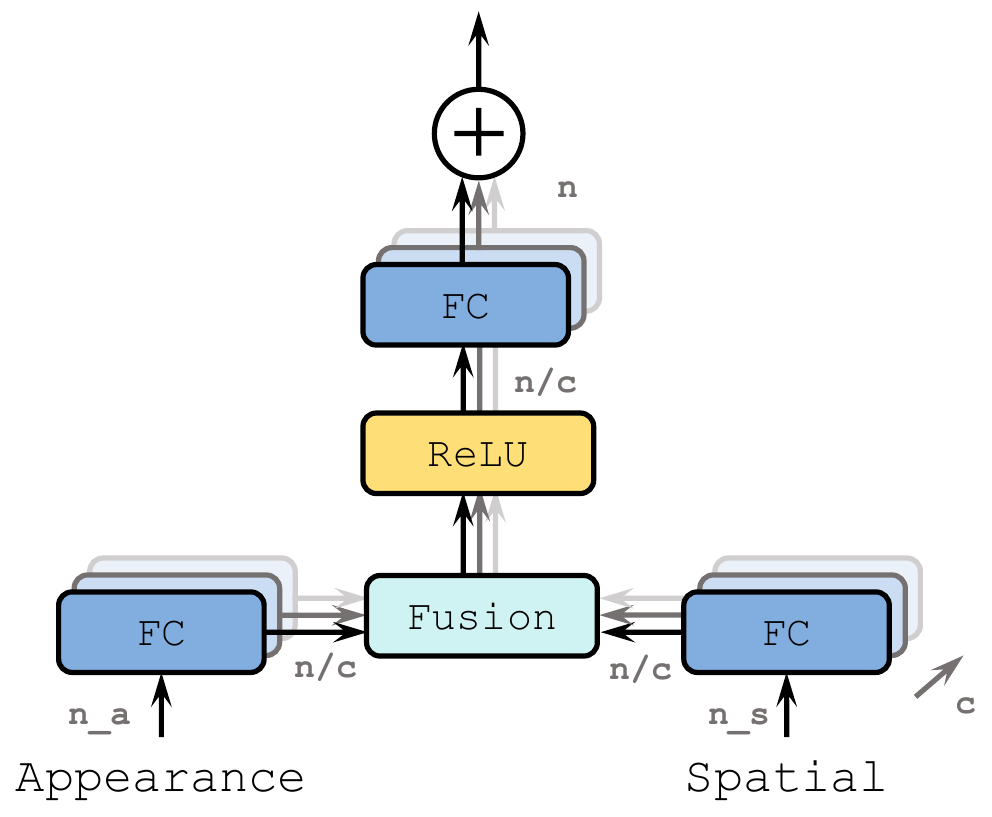}
	\caption{Structure of the multi-branch fusion module. The appearance and spatial features are mapped to \textit{c} subspaces, fused and mapped to an intermediate representation size. The outputs of different branches are aggregated by taking the sum. The input and output dimensions of each FC layer are marked in the diagram.}
	\label{fig:mbf}
\end{figure}

To increase the expressive power of the spatial conditioning, we use a multi-branch structure for modality fusion. We map the modalities to $c$ subspaces with reduced dimension, fuse the projections in each subspace, and then aggregate the outputs, as shown in Figure~\ref{fig:mbf}.
We refer to the proposed module as \textit{multi-branch fusion} (MBF). Following the nomenclature of Xie \etal~\cite{xie2017}, we refer to the number of homogeneous branches as the \textit{cardinality}.
Importantly, the number of parameters is independent of the cardinality by design, due to the subspace dimensionality reduction.
We define the message functions as
\begin{align}
   M_{\cO \rightarrow \cH}(\by_{j}^{t}, \bz_{ij}) &= \text{MBF}_{o}(\by_{j}^{t}, \bz_{ij}) \label{eq:msg_h}\\
   M_{\cH \rightarrow \cO}(\bx_{i}^{t}, \bz_{ij}) &= \text{MBF}_{h}(\bx_{i}^{t}, \bz_{ij}) \label{eq:msg_o}.
\end{align}
The two fusion modules have independent weights, allowing for anisotropic messages.

MBFs are also used to compute the adjacency with spatial conditioning, with an additional linear layer to map the output to a scalar. The pre-normalised adjacency is
\begin{equation}
   \label{eq:adj}
   \widetilde{\alpha}_{k} = \bw_{k}\transpose \:\sigma\left(\text{MBF}_{\alpha}(\bx_{i}^{t} \oplus \by_{j}^{t}, \bz_{ij})\right) + b_{k}
\end{equation}
where $\bw_{k} \in \reals^{n}$, $b_{k} \in \reals$ and \textit{k} is a linear index corresponding to a pair of $(i, j)$, that is, $k \in \{1,...,\vert \cH \times \cO \vert\}$. During message passing, the adjacency value $\alpha_{ij}^{r}$ is obtained by applying softmax to the entries sharing the same index $i$ (row normalisation). Similarly, $\alpha_{ij}^{c}$ is obtained via softmax while fixing $j$ (column normalisation).

After all iterations of message passing, we fuse the spatial features and the graph features prior to binary classification for each target class. The computation of classification scores has the same form as that of the adjacency matrix in (Eq.~\ref{eq:adj}), except with an additional sigmoid layer and that the output dimension is equal to the number of target classes. In fact, the adjacency can be interpreted as general interactiveness while the class probabilities are further conditioned on action types. For this reason, we use the same MBF module to compute the adjacency matrix and class probabilities.

\subsection{Contextual Cues}

As with most RoIPool-based feature extraction methods, the pooled information is local to a region. While this is reasonable for object detection, longer-range information about the context or even the global scene can be crucial for understanding human--object interactions. While Qi \etal~\cite{qi2018} used appearance features extracted from the minimum covering rectangle of the human and object boxes as edge features, our model uses spatial information as edge features. To compensate for the loss of contextual cues, we employ another MBF module to fuse the global features and the spatial features for each pair as $\text{MBF}_{g}(\bg, \bz_{ij})$, where $\bg$ represents the global features. These features are concatenated with the spatially conditioned graph features to give $\text{MBF}_{\alpha}(\bx_{i}^{T} \oplus \by_{j}^{T}, \bz_{ij}) \oplus \text{MBF}_{g}(\bg, \bz_{ij})$ for classification.

\subsection{Training and Inference}

For each image during training, we append the ground-truth boxes to the set of detections and assign them a score of one. We then remove detected boxes below a threshold score and apply non-maximum suppression. The $m$ highest scoring human and object boxes are then selected to initialise the bipartite graph. After message passing, we generate a set of human--object pairs from the graph, denoted by $\{q_{k}\}_{k=1}^{\scriptscriptstyle{\vert \cH \times \cO \vert}}$, where $q_{k}=(\bb_{i}^{h}, s_{i}^{h}, \bb_{j}^{o}, s_{j}^{o}, \widetilde{\bs}_{k})$.
The bounding boxes $\bb$ and object detection scores $s$ are obtained from the corresponding human and object nodes connected by edge $(i, j)$. The classification scores for all actions $\widetilde{\bs}_{k}$ are multiplied by the object detections scores.
In practice, however, because the object detection scores do not consider the interactiveness of object instances, they tend to be over-confident. As a result, we raise the object detection scores to the power of $\lambda$ during inference to counter this effect. The purpose of such operation is the same as the \textit{Low-grade Instance Suppression} function~\cite{li2019}. But we found raising the power works better for our model. The final scores are computed as
\begin{equation}
   \bs_{k} = \left(s_{i}^{h}\right)^{\lambda} \cdot \left( s_{j}^{o}\right)^{\lambda} \:\: \widetilde{\bs}_{k}.
\end{equation}

To associate the detected human--object pairs with the ground truth, the intersection-over-union is computed between each detected pair and ground-truth pair. Following previous practice~\cite{chao2018}, the IoU is computed for human and object boxes separately and taken as the minimum of the two. Detected pairs are considered to be positive when the IoU is above a designated threshold.

Due to the nature of proposal generation, there are overwhelmingly more negative examples than positive ones. In particular, the majority of examples are easy negatives. This inhibits the model from further improving on examples that are not well classified. To alleviate this issue, we adopt the focal loss~\cite{lin2017iccv} as a binary classification loss, given by
\begin{equation}
   \text{FL}(\hat{y}, y) =
   \left\{ \begin{array}{ll}
      - \beta (1-\hat{y})^{\gamma} \log(\hat{y}), & y = 1 \\
      - (1-\beta) \:\hat{y}^{\gamma} \log(1-\hat{y}), & y = 0
   \end{array} \right.
\end{equation}
where $\hat{y} \in [0, 1]$ is the final score of an example for a certain class, $y \in \{0, 1\}$ is the binary label, and $\beta\in [0, 1]$ and $\gamma \in \reals_{+}$ are hyper-parameters. In particular, $\beta$ is a balancing factor between positive and negative examples. With $\beta>0.5$, positive examples are assigned higher weights and vice versa. The parameter $\gamma$ attenuates the loss incurred on well-classified examples. This prevents the large number of easy negatives from dominating the gradient. However, suppressing easy negatives reduces the focal loss' magnitude \cite{lin2017iccv}, so normalisation is required. We extend Lin \etal's~\cite{lin2017iccv} proposal to binary classification by normalising the loss by the number of positive logits.

It is also important to restrict the output space to meaningful interactions. Denote the set of actions by $\cA$ and the subset of valid actions for a specific object type $o \in \cK$ by $\cA_{o}$. Then the interactions of interest are in the set $\cI = \cup_{o \in \cK} \:\cA_{o} \times \{o\}$, with $\cI \subseteq \cA \times \cK$. Following the practice of Gupta \etal~\cite{gupta2019}, we only compute the loss on the subset $\cA_{o}$ for each human--object pair, given the object type $o$. This removes predictions for non-existent interaction types, such as \textit{eating a car}, allowing the network to dedicate its parameters to learning meaningful interactions.

In the HICO-DET dataset~\cite{chao2018}, interactions of interest include those between two humans (\ie, a human may be an object and a subject in an HOI triplet). To capture such interactions, we construct bipartite graphs such that object nodes subsume human nodes, that is, object nodes are identical to the set of all detections. Human nodes representing the same instance across the bipartition are initialised to be the same, yet will diverge as message passing proceeds.

\section{Experiments}

\subsection{Dataset and Metric}

We evaluated our model on the HICO-DET~\cite{chao2018} and V-COCO~\cite{gupta2015} datasets.
HICO-DET contains $37\,633$ training and $9\,546$ test images with bounding box annotations, 80 object classes (identical to those in the MS COCO dataset~\cite{lin2014}), 117 action types, and 600 interaction types.
There are $117\,871$ annotated human--object pairs in the training set and $33\,405$ in the test set.
The distribution of pairs per interaction class is highly uneven, following a long tail distribution. In particular, there are 47 interaction categories with only one training example.

The evaluation metric is mean average precision (mAP). Detected human--object pairs are considered as positive when the IoU with any ground-truth pair is higher than 0.5. For multiple detected pairs associated with the same ground-truth instance, only the highest scoring pair is considered as positive. The computation of mAP follows the 11-point interpolation algorithm used in the Pascal VOC challenge~\cite{everingham2014}.
To capture the effectiveness of our model across interactions with different numbers of annotations, we follow previous practice~\cite{chao2018} and report results in three categories: full (all 600 interactions), rare (138 interactions with fewer than 10 training examples), and non-rare (462 interactions with 10 or more training examples).

V-COCO is a much smaller dataset with $2\,533$ images in the training set, $2\,867$ in the validation set and $4\,946$ in the test set. The dataset contains 26 different actions. We report our performance on this dataset for legacy reasons.

\subsection{Implementation Details}

We use Faster R-CNN~\cite{ren2015} with ResNet50-FPN~\cite{he2016, lin2017} pretrained on MS COCO~\cite{lin2014} to generate detections. For each image, we first filter out detections with scores lower than 0.2 and perform non-maximum suppression (NMS) with a threshold of 0.5. Afterwards, we extract the $m=15$ highest scoring human boxes, and the $m=15$ highest scoring object boxes. This gives us at most $15(30-1)=435$ box pairs, after removing pairs involving the same person twice. Inference follows the same setup, except that ground-truth detections are not used.

We use ResNet50-FPN ~\cite{he2016, lin2017} as the backbone for feature extraction. To utilise the feature pyramid, boxes are assigned to different pyramid levels based on their sizes~\cite{lin2017}. The pooled box features are mapped to 1024-dimensional vectors with a two-layer MLP. Similarly, the spatial features are mapped to the same dimension (1024) with a three-layer MLP. For the MBF module, we use $c=16$ and $n=1024$. We use $T=2$ iterations of message passing for all models unless otherwise specified. To counter the over-confidence in object scores, we set $\lambda=2.8$ during inference while keeping $\lambda=1.0$ during training. Lastly, for the focal loss, we set $\beta=0.5$ and $\gamma=0.2$. All hyper-parameters are selected using cross-validation.

We adopt an image-centric training strategy~\cite{girshick2015} with slight modifications. Input images are normalised and resized such that the shorter edge is 800 pixels. Bounding boxes are then resized accordingly. Afterwards, images are batched with zero padding. To train the model, we use AdamW~\cite{ilya2018} as the optimiser, with a momentum of 0.9 and weight decay of $10^{-4}$. We use an initial learning rate of $10^{-5}$ for the backbone and $10^{-4}$ for the rest of the network. The learning rates are dropped by a magnitude at the sixth epoch. All models are trained for 10 epochs on 8 GeForce GTX TITAN X devices, with an effective batch size of 32.

\subsection{Comparison with State-of-the-Art}

\begin{table}[!t]\small
	\caption{HOI detection performance (mAP$\times100$) on the HICO-DET~\cite{chao2018} test set under the default setting. See appendix for the known object setting. The most competitive method in each category is in bold, while the second best is underlined.}
	\label{tab:results_hico}
	\setlength{\tabcolsep}{2pt} 
	\begin{tabularx}{\linewidth}{l l C C c}
		\toprule
		\textbf{Method} & \textbf{Backbone} & \textbf{Full} & \textbf{Rare} & \textbf{Non-rare} \\
		\midrule
		\multicolumn{5}{c}{\sc Detector pre-trained on MS COCO} \\
		HO-RCNN~\cite{chao2018} & CaffeNet & 7.81 & 5.37 & 8.54 \\
		InteractNet~\cite{gkioxari2018} & ResNet-50-FPN & 9.94 & 7.16 & 10.77 \\
		GPNN~\cite{qi2018} & ResNet-101 & 13.11 & 9.34 & 14.23 \\
		iCAN~\cite{gao2018} & ResNet-50 & 14.84 & 10.45 & 16.15 \\
		Bansal \etal~\cite{bansal2020} & ResNet-101 & 16.96 & 11.73 & 18.52 \\
		TIN~\cite{li2019} & ResNet-50 & 17.03 & 13.42 & 18.11 \\
		Gupta \etal~\cite{gupta2019} & ResNet-152 & 17.18 & 12.17 & 18.68 \\
      RPNN~\cite{zhou2019} &  ResNet-50 & 17.35 & 12.78 & 18.71 \\
		Wang \etal~\cite{wang2020} & ResNet-50-FPN & 17.57 & 16.85 & 17.78 \\
		DRG~\cite{gao2020} & ResNet-50-FPN & 19.26 & 17.74 & 19.71 \\
		Peyre \etal~\cite{peyre2019} & ResNet-50-FPN & 19.40 & 14.63 & 20.87 \\
      VCL~\cite{hou2020} & ResNet50 & 19.43 & 16.55 & 20.29 \\
		VSGNet~\cite{ulutan2020} & ResNet-152 & 19.80 & 16.05 & 20.91 \\
      IDN~\cite{li2020} & ResNet50 & \textbf{23.36} & \textbf{22.47} & \textbf{23.63} \\
		Ours & ResNet-50-FPN & \underline{21.85} & \underline{18.11} & \underline{22.97} \\
		\midrule
		\multicolumn{5}{c}{\sc Detector fine-tuned on HICO-DET} \\
		PPDM~\cite{liao2020} & Hourglass-104 & 21.73 & 13.78 & 24.10 \\
		Bansal \etal~\cite{bansal2020} & ResNet-101 & 21.96 & 16.43 & 23.63 \\
      VCL~\cite{hou2020} & ResNet50 & 23.63 & 17.21 & 25.55 \\
		DRG~\cite{gao2020} & ResNet-50-FPN & 24.53 & 19.47 & 26.04 \\
      IDN~\cite{li2020} & ResNet50 & \underline{26.29} & \underline{22.61} & \underline{27.39} \\
		Ours & ResNet-50-FPN & \textbf{31.33} & \textbf{24.72} & \textbf{33.31} \\
		\midrule
		\multicolumn{5}{c}{\sc Oracle Detector} \\
		iCAN~\cite{gao2018} & ResNet-50 &  33.38 & 21.43 & 36.95 \\
      TIN~\cite{li2019} & ResNet50 & 34.26 & 22.90 & 37.65 \\
		Peyre \etal~\cite{peyre2019} & ResNet-50-FPN & 34.35 & 27.57 & 36.38 \\
      IDN~\cite{li2020} & ResNet50 & \underline{43.98} & \underline{40.27} & \underline{45.09} \\
		Ours & ResNet-50-FPN & \textbf{51.53} & \textbf{41.01} & \textbf{54.67} \\
		\bottomrule
	\end{tabularx}
\end{table}

\begin{table}[!t]\small
	\caption{Performance (mAP$\times100$) on the V-COCO~\cite{gupta2015} test set. The most competitive method in each category is in bold, while the second best is underlined. $^\star$Using a fine-tuned detector.}
	\label{tab:results_vcoco}
	\setlength{\tabcolsep}{2pt} 
	\begin{tabularx}{\linewidth}{l l C c}
		\toprule
		\textbf{Method} & \textbf{Backbone} & \textbf{Scenario 1} & \textbf{Scenario 2} \\
		\midrule
		InteractNet~\cite{gkioxari2018} & ResNet-50-FPN & 40.0 & -- \\
		GPNN~\cite{qi2018} & ResNet-101 & 44.0 & -- \\
		iCAN~\cite{gao2018} & ResNet-50 & 45.3 & 52.4 \\
		TIN~\cite{li2019} & ResNet-50 & 47.8 & 54.2 \\
		DRG~\cite{gao2020} & ResNet-50-FPN & 51.0 & -- \\
		VSGNet~\cite{ulutan2020} & ResNet-152 & 51.8 & 57.0 \\
		Wang \etal~\cite{wang2020} & ResNet-50-FPN & 52.7 & -- \\
      IDN~\cite{li2020} & ResNet50 & \textbf{53.3} & \textbf{60.3} \\
		Ours & ResNet-50-FPN & \underline{53.0} & \underline{58.2} \\
		\midrule
		Ours$^\star$ & ResNet-50-FPN & \textbf{54.2} & \textbf{60.9} \\
		\bottomrule
	\end{tabularx}
\end{table}

\begin{figure}[!t]\centering
	\includegraphics[width=0.49\linewidth]{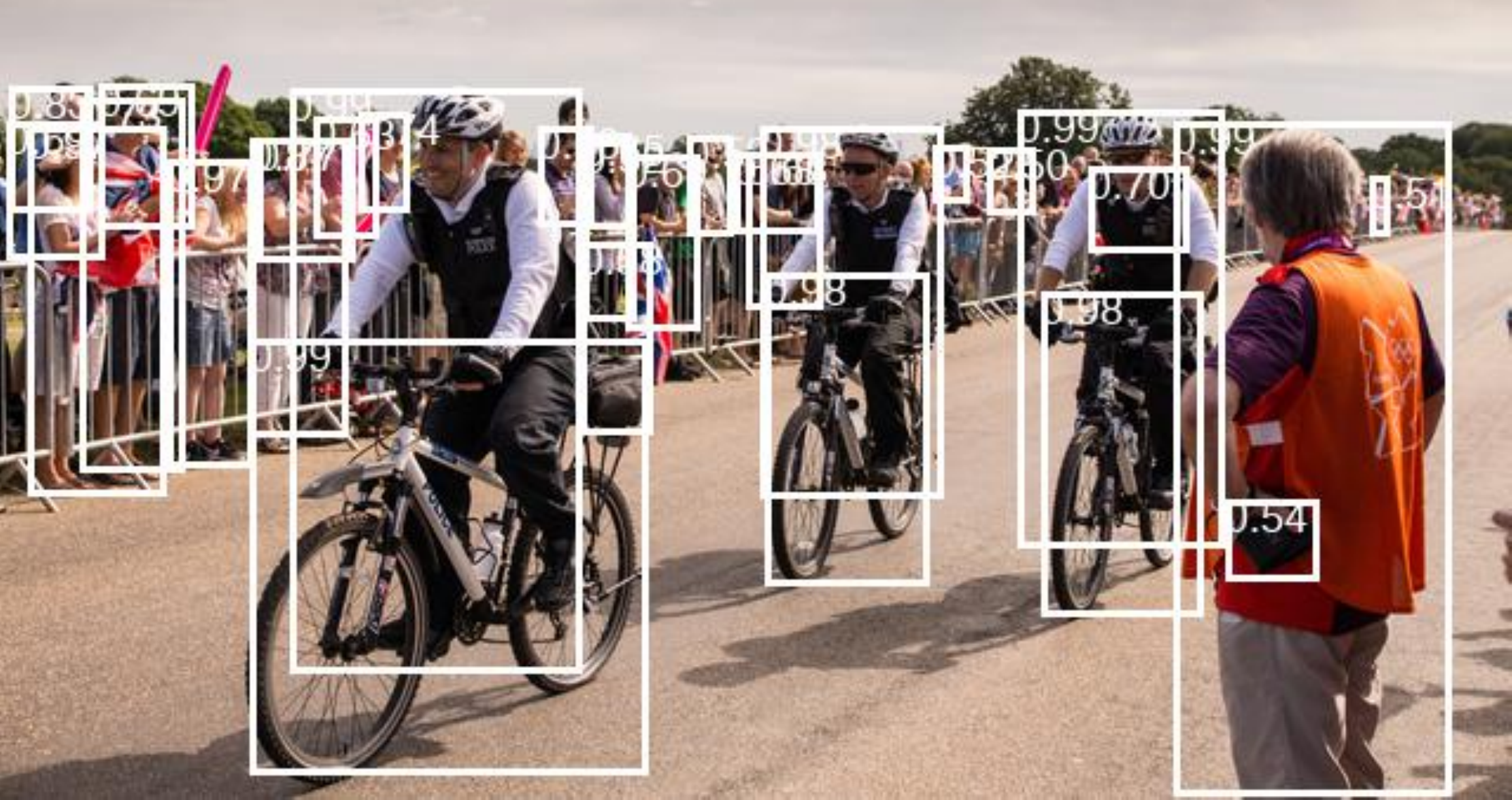}\hfill
	\includegraphics[width=0.49\linewidth]{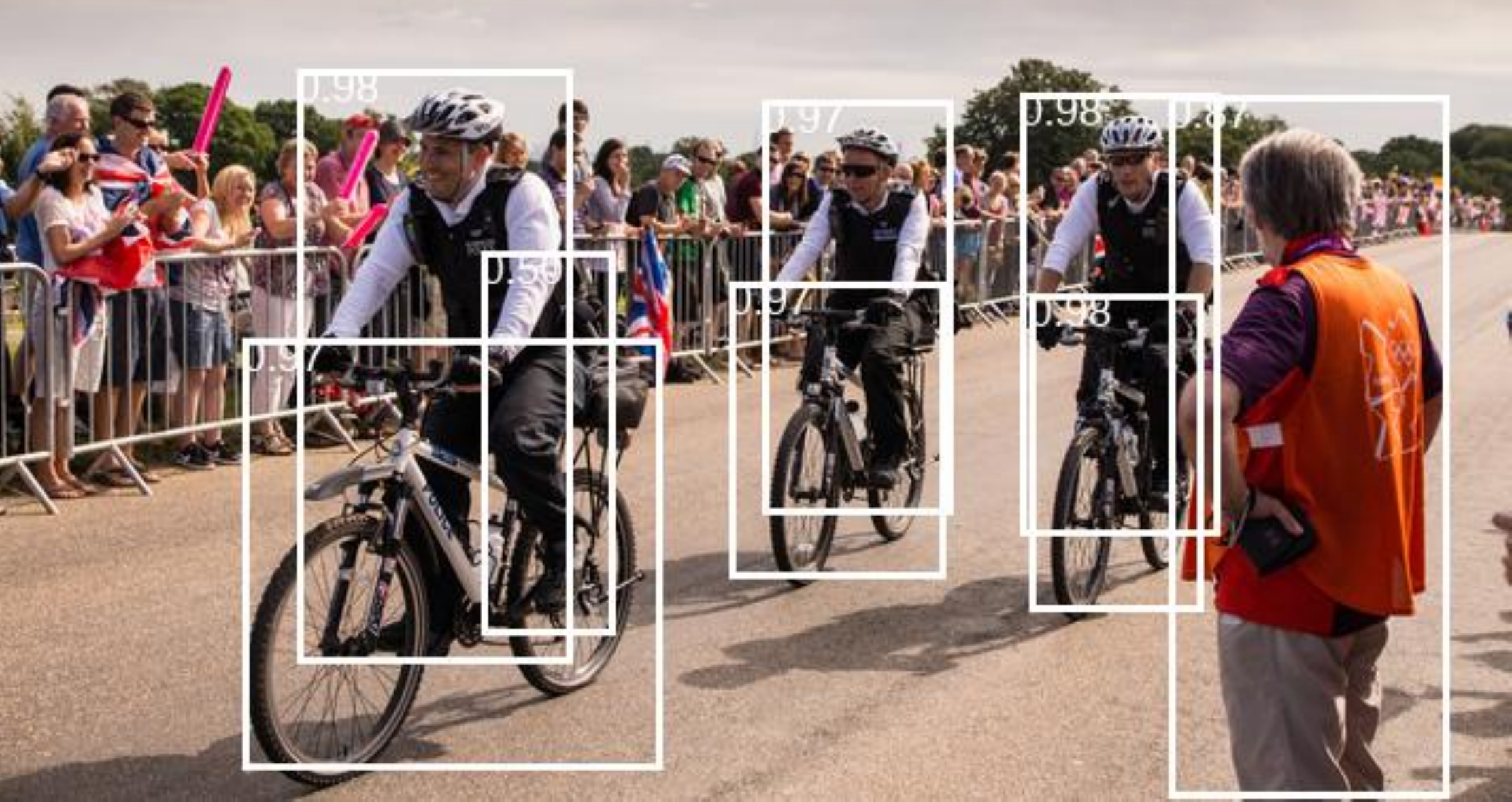}
	\caption{Object detections from the pre-trained MS COCO model (left) compared to the fine-tuned HICO-DET model (right). Boxes with scores higher than 0.5 are displayed. The fine-tuned detector suppresses objects that are less likely to be engaged in interactions.}
	\label{fig:detector}
\end{figure}

Quantitative results on the HICO-DET~\cite{chao2018} test set are shown in Table~\ref{tab:results_hico}. We report the performance of our model with three different detectors: one pre-trained on the MS COCO dataset~\cite{lin2014}, one fine-tuned on the HICO-DET dataset as provided by Gao \etal~\cite{gao2020}, and an oracle supplying the ground truth detections. We achieve competitive performance when using the COCO pre-trained detector, but significantly outperform state-of-the-art when using the higher-quality fine-tuned detections, a 20\% relative improvement. In particular, we outperform the next best method IDN~\cite{li2020} by 5~mAP, despite slightly underperforming that method when using the pre-trained detections. This suggests that our graph neural network can better exploit the high-quality detections. This is supported by the results for the oracle detector, where we outperform the next best method by 7.5~mAP. We show an example of the different detector outputs in Figure~\ref{fig:detector}. Less salient people and objects are suppressed in the fine-tuned detector, making the spatial information more discriminative. We also report on V-COCO~\cite{gupta2015} test set, as shown in Table~\ref{tab:results_vcoco}. Our model achieves competitive performance using a pre-trained detector and receives consistent gains from a fine-tuned detector.

\subsection{Contribution of Different Modalities}

\begin{table}[h!]\small
   \caption{Difference in performance between models with appearance and spatial features (Ours) and with only appearance features (baseline), as detection quality increases to the right.}
   \label{tab:perf_diff}
	\setlength{\tabcolsep}{4pt} 
	\begin{tabularx}{\linewidth}{l  C C C}
		\toprule
      Detector & \textbf{COCO} & \textbf{HICO-DET} & \textbf{Oracle} \\
      \midrule
		Performance $\Delta$ & +1.93 & +2.90 & +4.36 \\
		\bottomrule
	\end{tabularx}
\end{table}

\begin{figure}[!t]\centering
	\includegraphics[width=0.45\linewidth]{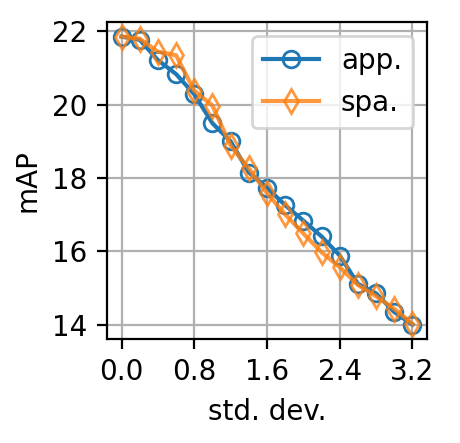}
	\includegraphics[width=0.45\linewidth]{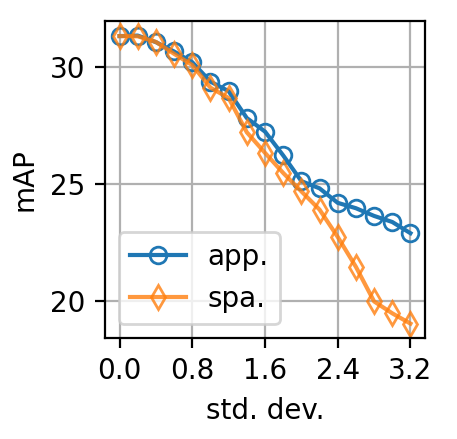}
	\caption{Model performance under different levels of corruption in appearance and spatial modalities, using the pre-trained detector (left) and the fine-tuned detector (right).}
	\label{fig:corruption}
\end{figure}

Notably, our model is able to gain nearly 9~mAP by using a fine-tuned detector and a further 20~mAP by using an oracle detector that supplies ground-truth detections, which is much higher than what previous methods gain. Due to the use of spatial conditioning in our model, we hypothesise that as detection quality improves, spatial information plays a more significant role in the disambiguation of interactions, while coarse appearance features contribute relatively less. This is supported by evidence in Table~\ref{tab:perf_diff}, where we show that the performance difference between the baseline model and our full model increases as detection quality improves. To investigate this hypothesis, we add Gaussian noise with zero mean and variable standard deviation to the appearance and spatial features separately, and observe how corruption in different modalities damages the performance. As shown in Figure~\ref{fig:corruption}, when using the pre-trained detector, noise in the appearance and spatial features has an approximately equal effect on performance. However, with the fine-tuned detector, noisy spatial features have a much larger impact. We conclude that spatial information contributes relatively more to performance as the detection quality improves.

\subsection{Ablation Studies}
\label{sec:ablation}

\begin{table}[t]\small
   \caption{Ablating the addition of spatial conditioning at different stages of the model on the HICO-DET dataset (mAP$\times100$).}
   \label{tab:fusion}
	\setlength{\tabcolsep}{4pt} 
	\begin{tabularx}{\linewidth}{l C c}
		\toprule
      \textbf{Stage} & \textbf{COCO Detector} & \textbf{HICO-DET Detector} \\
      \midrule
		None & 19.92 & 28.43 \\
		Adjacency & 20.56 & 29.48 \\
		Messages & 20.79 & 30.06 \\
		Global features & 20.44 & 29.51 \\
		Refinement & 21.03 & 30.11 \\
		All (Ours) & \textbf{21.85} & \textbf{31.33}\\
		\bottomrule
   \end{tabularx}
\end{table}

\begin{table}[t]\small
   \caption{Ablating the multi-branch fusion design choices, including the binary operation and the cardinality ($c$).}
   \label{tab:cardinality}
	\setlength{\tabcolsep}{4pt} 
	\begin{tabularx}{\linewidth}{l C c}
		\toprule
      \textbf{Design Choice} & \textbf{COCO Detector} & \textbf{HICO-DET Detector} \\
      \midrule
		Product ($c=1$) & 21.18 & 30.75 \\
      Sum ($c=1$) & \textbf{21.35} & \textbf{30.87} \\
      Concat. ($c=1$) & 21.02 & 30.66 \\
      \midrule
		Product ($c=16$) & \textbf{21.85} & 31.33 \\
      Sum ($c=16$) & 21.81 & 31.07 \\
      Concat. ($c=16$) & 21.67 & \textbf{31.65} \\
		\bottomrule
	\end{tabularx}
\end{table}

\begin{table}[t]\small
   \caption{Varying the number of message passing iterations ($T$).}
   \label{tab:iterations}
	\setlength{\tabcolsep}{4pt} 
	\begin{tabularx}{\linewidth}{l C c}
		\toprule
      \textbf{Model} & \textbf{COCO Detector} & \textbf{HICO-DET Detector} \\
      \midrule
		Ours ($T=0$) & 20.05 & 28.86 \\
		Ours ($T=1$) & 20.70 & 30.99 \\
		Ours ($T=2$) & \textbf{21.85} & 31.33 \\
		Ours ($T=3$) & 21.72 & \textbf{31.78} \\
		\bottomrule
	\end{tabularx}
\end{table}

We conducted a series of ablation studies to validate our design choices. Our baseline is a bipartite graph with appearance features only. Specifically, the message sent from a node is computed from its appearance encodings using a linear layer. The adjacency and class probabilities are computed from the concatenated node encodings of a human--object pair using an MLP, and the computation of classification scores shares weights with that of adjacency until the logistic layer.
We first investigate the importance of spatial conditioning at different stages in our model: for computing the adjacencies, messages, global features, and refined graph features. As shown in Table~\ref{tab:fusion}, every stage improves over the baseline, and they combine together to achieve the best performance.
We next demonstrate the impact of different design choices for multi-branch fusion, including the choice of fusion methods and the number of branches (cardinality). As shown in Table~\ref{tab:cardinality}, the performance improves with higher cardinality.
We also show that the performance is insensitive to the choice of binary fusion operation, with our choice (elementwise product) being comparable to the elementwise sum and concatenation operations.
Lastly, we show how the number of message passing iterations at test time affects the results. As shown in Table~\ref{tab:iterations}, message passing is clearly helpful for this problem, while an additional iteration further improves the results significantly.

\subsection{Qualitative Results}

\begin{figure}[!t]\centering
   \begin{subfigure}[t]{\linewidth}
      \centering
	   \includegraphics[width=0.8\linewidth]{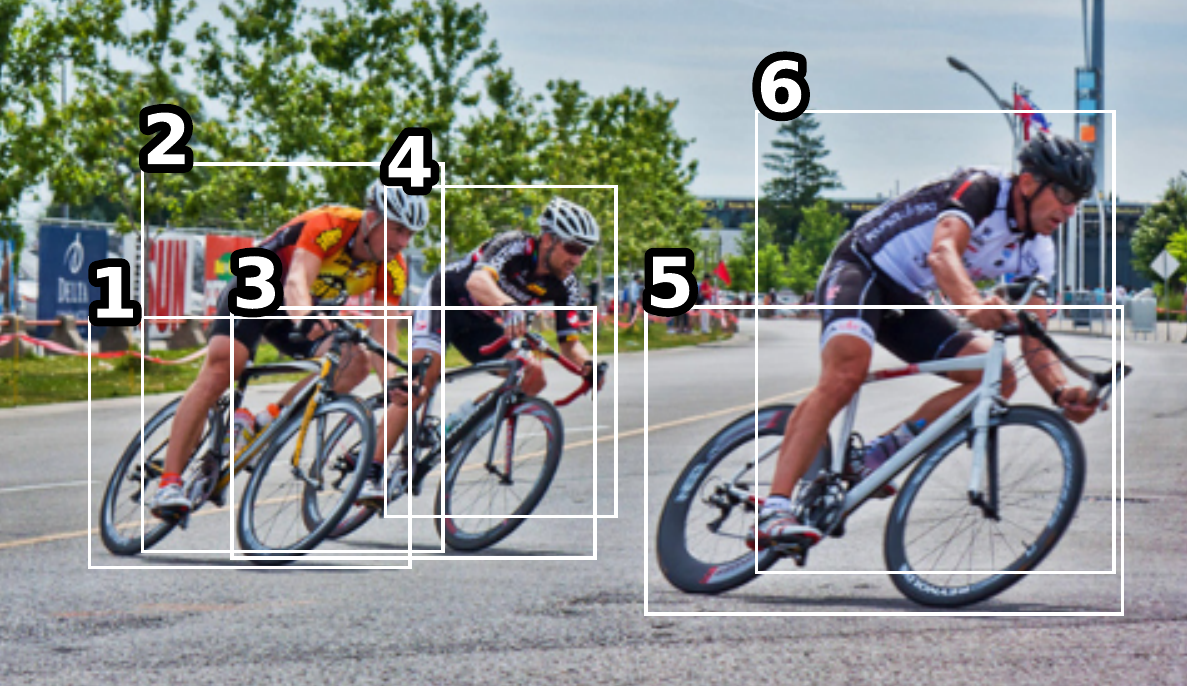}%
	   \vspace{-2pt}
      \caption{Interaction: \textit{riding a bike}}
      \label{fig:qual_1}
   \end{subfigure}%
	\vfill
   \begin{subfigure}[t]{\linewidth}
      \centering
	   \includegraphics[width=0.8\linewidth]{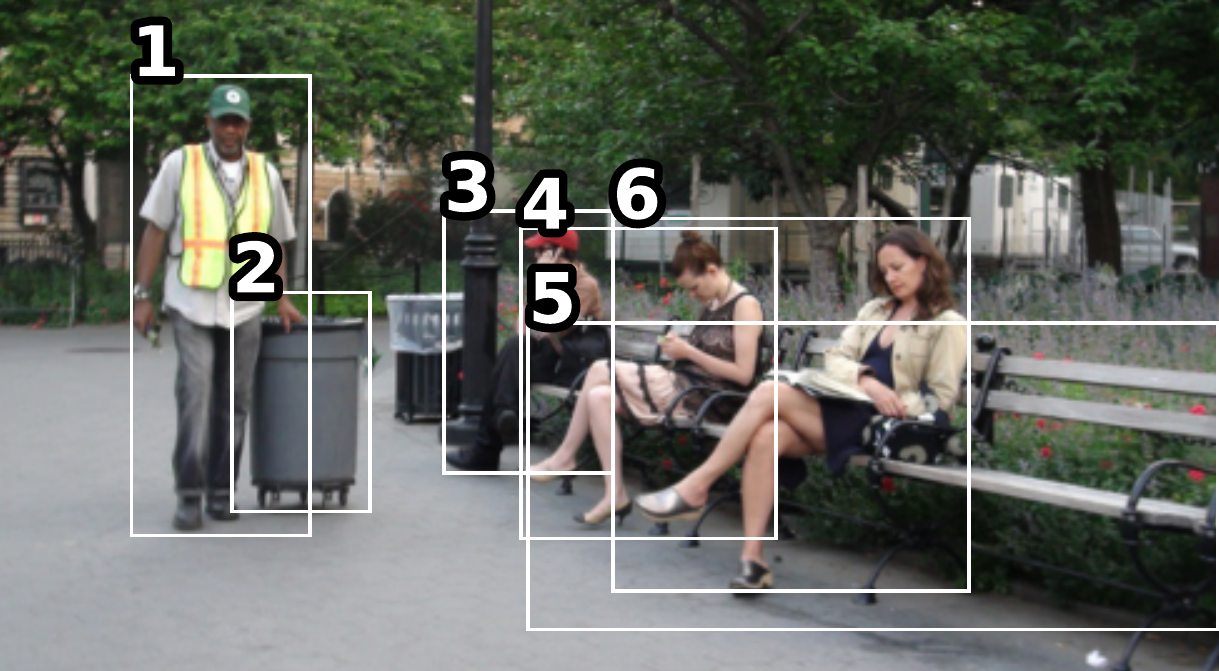}%
	   \vspace{-2pt}
      \caption{Interaction: \textit{sitting on a bench}}
      \label{fig:qual_2}
   \end{subfigure}%
	\vspace{4pt}
	\caption{Qualitative results with success and failure cases for our model. The scores corresponding to~(\subref{fig:qual_1}) are in Table~\ref{tab:qual_1}, and the scores for~(\subref{fig:qual_2}) are in Table~\ref{tab:qual_2}.}
	\label{fig:qualitative}
	\vspace{-3pt}
\end{figure}

We show qualitative results of our model in Figure~\ref{fig:qualitative}. In Figure~\ref{fig:qual_1}, the ground-truth interaction is \textit{riding a bike}. As shown in Table~\ref{tab:qual_1}, positive human--bike pairs $(1, 2)$, $(3, 4)$ and $(5, 6)$ have the highest scores. However, the network also assigns the negative pair $(2, 3)$ a relatively high score. This is due to the spatial proximity and visual similarity between bike ($3$) and the correct bike instance ($1$), which can confuse our model. Another example is given in Figure~\ref{fig:qual_2}, where the true interaction is \textit{sitting on a bench}. As shown in Table~\ref{tab:qual_2}, our model assigns high scores to all correct human--bench pairs and suppresses the non-interactive pair $(1, 5)$.

\begin{table}[h!]\small
   \caption{Scores for the interaction \textit{riding a bike} in Figure~\ref{fig:qual_1}.}
   \label{tab:qual_1}
	\setlength{\tabcolsep}{4pt} 
	\begin{tabularx}{\linewidth}{c | C C C}
		\toprule
      Instance index & \textbf{2} & \textbf{4} & \textbf{6} \\
      \midrule
		\textbf{1} & \textbf{0.5742} & 0.0027 & 0.0000 \\
		\textbf{3} & 0.4617 & \textbf{0.4735} & 0.0002\\
		\textbf{5} & 0.0006 & 0.0008 & \textbf{0.7899}\\
		\bottomrule
	\end{tabularx}
\end{table}

\begin{table}[h!]\small
	\vspace{-15pt}
   \caption{Scores for the interaction \textit{sitting on a bench} in Figure~\ref{fig:qual_2}.}
   \label{tab:qual_2}
	\setlength{\tabcolsep}{4pt} 
	\begin{tabularx}{\linewidth}{c | C C C C}
		\toprule
      Instance index & \textbf{1} & \textbf{3} & \textbf{4} & \textbf{6}  \\
      \midrule
		\textbf{5} & 0.0011 & 0.3890 & 0.4049 & \textbf{0.6056} \\
		\bottomrule
	\end{tabularx}
\end{table}

\section{Conclusion}

In this paper, we have proposed a spatially conditioned graph neural network for detecting human--object interactions. To perform spatial conditioning, we applied a multi-branch fusion mechanism that modulates the appearance features with the spatial configuration of the human--object pairs. We use this mechanism consistently for computing adjacency, messages and refined graph features, and show that our model outperforms the state-of-the-art by a considerable margin with fine-tuned detections.
We also show that the margin of improvement increases with the detection quality, allowing our model to most effectively exploit advances in object detector research.

\section*{Acknowledgements}
This research is funded in part by the ARC Centre of Excellence for Robotic Vision (CE140100016) and Continental AG (D.C.).

{\small
\bibliographystyle{ieee_fullname}
\bibliography{egbib}
}

\appendix

\section{Known Object Setting for HICO-DET}

While the default setting for HICO-DET~\cite{chao2018} has been the more popular evaluation protocol, there is an additional less frequently reported known object setting, where the object types of ground truth interactions in images are considered known, thus automatically removing predicted interactive pairs with other object types. For interested readers, we provide the performance of our model in comparison with other methods under the known object setting in Table~\ref{tab:sup_results_hico}.

\begin{table}[!h]\small
	\caption{HOI detection performance (mAP$\times100$) on the HICO-DET~\cite{chao2018} test set under the known object setting. The most competitive method in each category is in bold, while the second best is underlined.}
	\label{tab:sup_results_hico}
	\setlength{\tabcolsep}{2pt} 
	\begin{tabularx}{\linewidth}{l l C C c}
		\toprule
		\textbf{Method} & \textbf{Backbone} & \textbf{Full} & \textbf{Rare} & \textbf{Non-rare} \\
		\midrule
		\multicolumn{5}{c}{\sc Detector pre-trained on MS COCO} \\
		HO-RCNN~\cite{chao2018} & CaffeNet & 10.41 & 8.94 & 10.85 \\
		iCAN~\cite{gao2018} & ResNet-50 & 16.26 & 11.33 & 17.73 \\
		TIN~\cite{li2019} & ResNet-50 & 19.17 & 15.51 & 20.26 \\
		DRG~\cite{gao2020} & ResNet-50-FPN & 23.40 & 21.75 & 23.89 \\
      	VCL~\cite{hou2020} & ResNet50 & 22.00 & 19.09 & 22.87 \\
      	IDN~\cite{li2020} & ResNet50 & \textbf{26.43} & \textbf{25.01} & \textbf{26.85} \\
		Ours & ResNet-50-FPN & \underline{25.53} & \underline{21.79} & \underline{26.64} \\
		\midrule
		\multicolumn{5}{c}{\sc Detector fine-tuned on HICO-DET} \\
		PPDM~\cite{liao2020} & Hourglass-104 & 24.58 & 16.65 & 26.84 \\
      	VCL~\cite{hou2020} & ResNet50 & 25.98 & 19.12 & 28.03 \\
		DRG~\cite{gao2020} & ResNet-50-FPN & 27.98 & 23.11 & \underline{29.43} \\
      	IDN~\cite{li2020} & ResNet50 & \underline{28.24} & \underline{24.47} & 29.37 \\
		Ours & ResNet-50-FPN & \textbf{34.37} & \textbf{27.18} & \textbf{36.52} \\
		\midrule
		\multicolumn{5}{c}{\sc Oracle Detector} \\
		Ours & ResNet-50-FPN & \textbf{51.75} & \textbf{41.40} & \textbf{54.84} \\
		\bottomrule
	\end{tabularx}
\end{table}

\section{Additional Qualitative Results}

\begin{figure}[t!]\centering
	\begin{subfigure}[t]{\linewidth}
	   \centering
		\includegraphics[width=0.8\linewidth,keepaspectratio]{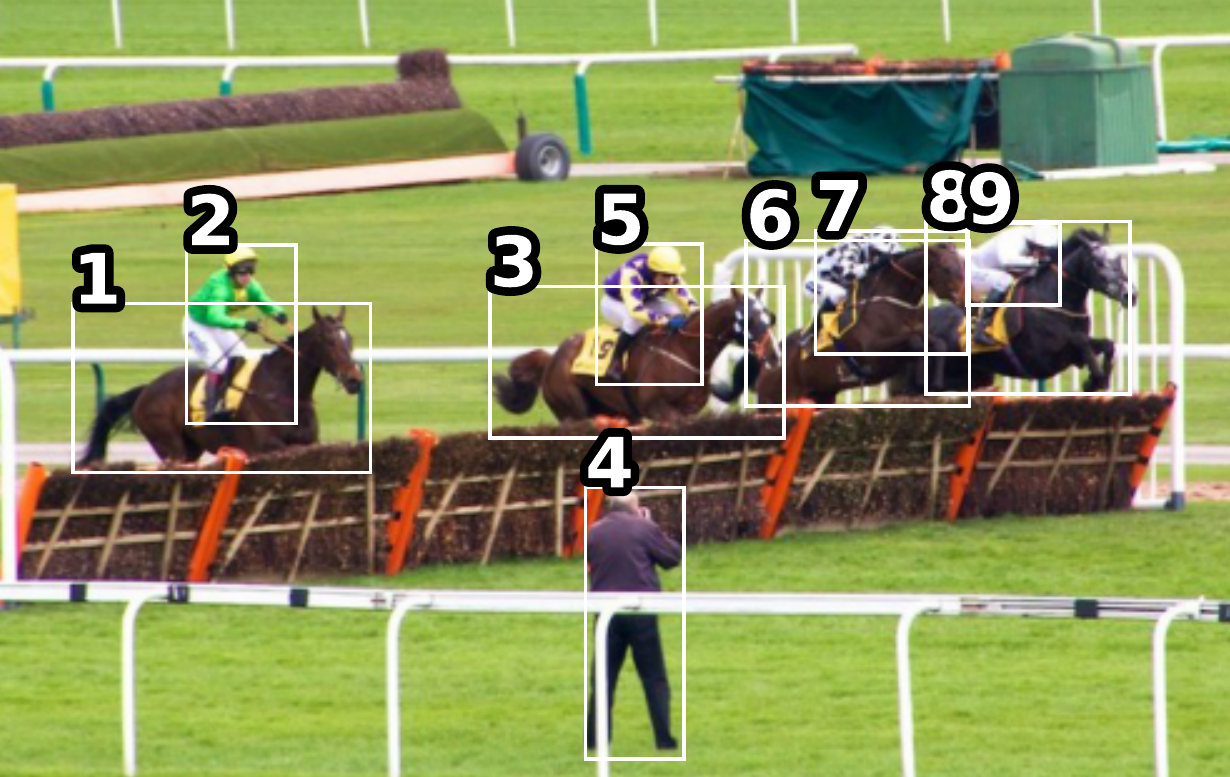}%
	   \caption{Interaction: \textit{racing a horse}}
	   \label{fig:sup_qual_1}
	\end{subfigure}
	\begin{subfigure}[t]{\linewidth}
		\centering
		 \includegraphics[width=0.8\linewidth,keepaspectratio]{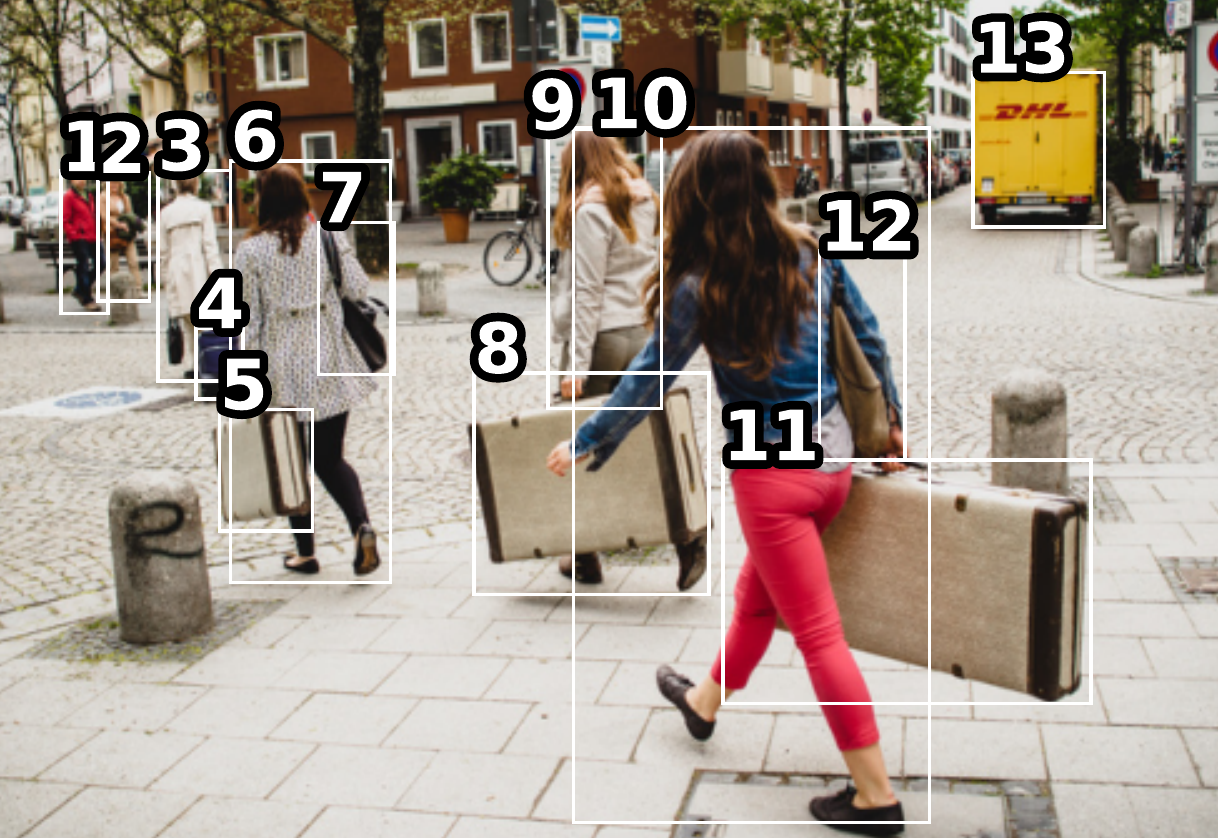}%
		\caption{Interaction: \textit{carrying a suitcase}}
		\label{fig:sup_qual_2}
	 \end{subfigure}
	 \newline
	 \caption{Qualitative results. The scores corresponding to~(\subref{fig:sup_qual_1}) are shown in Table~\ref{tab:sup_qual_1} and the scores corresponding to~(\subref{fig:sup_qual_2}) are shown in Table~\ref{tab:sup_qual_2}.}
	 \label{fig:sup_qualitative}
 \end{figure}

 \begin{table}[h!]\small
	\caption{Scores for the interaction \textit{racing a horse} in Figure~\ref{fig:sup_qual_1}. Each column corresponds to pairs with the same human instance. Each row corresponds to pairs with the same horse instance.}
	\label{tab:sup_qual_1}
	 \setlength{\tabcolsep}{2pt} 
	 \begin{tabularx}{\linewidth}{c | C C C C C}
		 \toprule
	   Instance index & \textbf{2} & \textbf{4} & \textbf{5} & \textbf{7} & \textbf{9} \\
	   \midrule
		 \textbf{1} & \textbf{0.2031} & 0.0000 & 0.0000 & 0.0000 & 0.0000 \\
		 \textbf{3} & 0.0000 & 0.0000 & \textbf{0.5913} & 0.0002 & 0.0000 \\
		 \textbf{6} & 0.0000 & 0.0000 & 0.0013 & \textbf{0.0178} & 0.0030 \\
		 \textbf{8} & 0.0000 & 0.0000 & 0.0001 & 0.0034 & \textbf{0.1412} \\
		 \bottomrule
	 \end{tabularx}
 \end{table}

 \begin{table}[h!]\small
	\caption{Scores for the interaction \textit{carrying a suitcase} in Figure~\ref{fig:sup_qual_2}. Each column corresponds to pairs with the same human instance. Each row corresponds to pairs with the same suitcase instance. Missing indices correspond to detections other than suitcases.}
	\label{tab:sup_qual_2}
	 \setlength{\tabcolsep}{2pt} 
	 \begin{tabularx}{\linewidth}{c | C C C C C C}
		 \toprule
	   Instance index & \textbf{1} & \textbf{2} & \textbf{3} & \textbf{6} & \textbf{9}  & \textbf{10} \\
	   \midrule
		 \textbf{4} & 0.0000 & 0.0000 & \textbf{0.0391} & 0.0021 & 0.0000 & 0.0000 \\
		 \textbf{5} & 0.0000 & 0.0000 & 0.0000 & \textbf{0.1278} & 0.0000 & 0.0004 \\
		 \textbf{8} & 0.0000 & 0.0000 & 0.0000 & 0.0178 & \textbf{0.2791} & \underline{0.1098} \\
		 \textbf{11} & 0.0000 & 0.0000 & 0.0000 & 0.0003 & 0.0000 & \textbf{0.3858} \\
		 \bottomrule
	 \end{tabularx}
 \end{table}

 We show more qualitative results to demonstrate the strength of our model in Figure~\ref{fig:sup_qualitative}. We intentionally select images that have many human instances and multiple human--object pairs of the same interaction. In Figure~\ref{fig:sup_qual_1}, there are 20 combinatorial human--horse pairs, with 4 of them being interactive. As shown in Table~\ref{tab:sup_qual_1}, our model is able to assign highest scores to all four interactive pairs and suppress all non-interactive pairs. However, we do notice that small and clustered boxes can reduce the confidence of our model, e.g. person ($7$) and horse ($6$). This issue can also be seen in Figure~\ref{fig:sup_qual_2} and Table~\ref{tab:sup_qual_2}. Our model is able to find the correct human--suitcase pairs ($10, 11$), ($9, 8$), ($6, 5$) and predict high scores for them. Yet the positive pair ($3, 4$) receives a very low score due to the size of the bounding boxes and less confident object detection scores. We also notice that person ($10$) and suitcase ($8$) receive a fairly high score for \textit{carrying a suitcase}. This is due to the close relative location between the pair and a plausible gesture from the person. In such scenarios, access to the depth information could be helpful.

 \begin{figure}[h!]\centering
	\begin{subfigure}[t]{.49\linewidth}
	   \centering
		\includegraphics[height=0.85\linewidth,keepaspectratio]{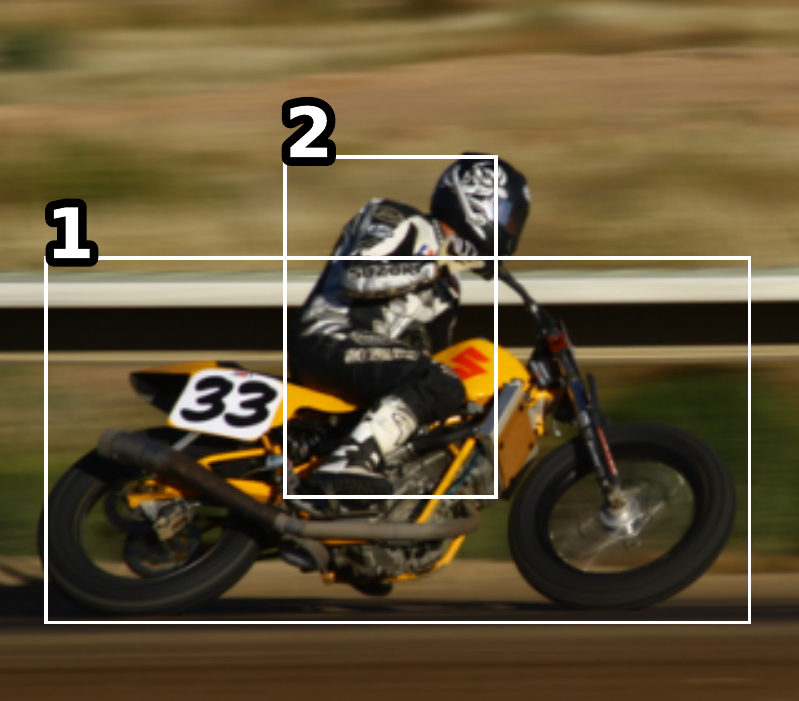}%
	   \caption{Interaction: \textit{racing a motocycle}}
	   \label{fig:sup_qual_3}
	\end{subfigure}
	\begin{subfigure}[t]{.49\linewidth}
		\centering
		 \includegraphics[height=0.85\linewidth,keepaspectratio]{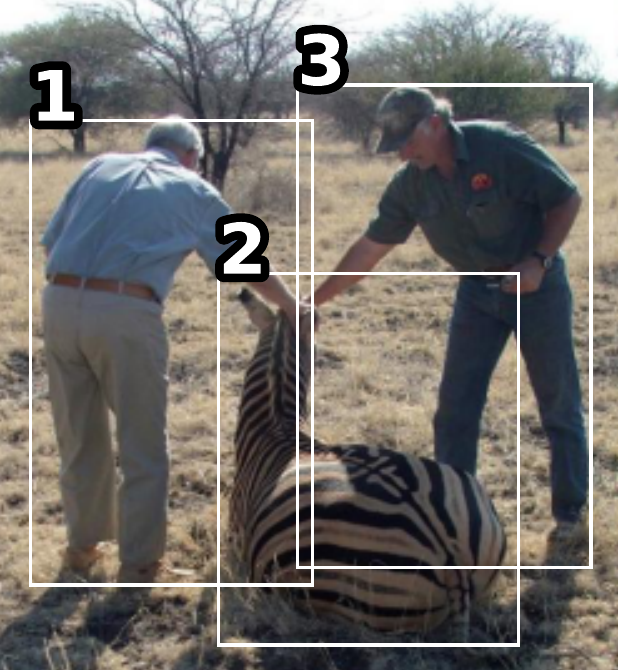}%
		\caption{Interaction: \textit{petting a zebra}}
		\label{fig:sup_qual_4}
	 \end{subfigure}
	 \newline
	 \caption{Qualitative results where images contain a small number of clean human and object instances.}
	 \label{fig:sup_qualitative_2}
 \end{figure}

 We also show some qualitative results where our model does not improve upon previous methods in Figure~\ref{fig:sup_qualitative_2}. For examples such as in Figure~\ref{fig:sup_qual_3}, where there is only one human--object pair, our graphical model is not particularly superior as there are only one human and object node each passing messages between each other. And in Figure~\ref{fig:sup_qual_4}, when both human--zebra pairs are in fact interactive under the interaction \textit{petting a zebra}, we found that the baseline model with appearance only is also able to correctly assign high scores to both pairs, as shown in Table~\ref{tab:sup_qual_4}.

 \begin{table}[h!]\small
	\caption{Scores for the interaction \textit{petting a zebra} in Figure~\ref{fig:sup_qual_4}}
	\label{tab:sup_qual_4}
	 \setlength{\tabcolsep}{2pt} 
	 \begin{tabularx}{\linewidth}{c | C C }
		 \toprule
	   Human--zebra pairs & Scores (baseline) & Scores (ours) \\
	   \midrule
		 (\textbf{1, 2}) & 0.6782 & 0.7019 \\
		 (\textbf{1, 3}) & 0.6945 & 0.6799 \\
		 \bottomrule
	 \end{tabularx}
 \end{table}

 To sum up, we found that our graphical model with spatial conditioning is more competitive on images with large number of human and object instances, particularly when there are multiple ground truth pairs of the same interaction, but does not improve upon previous methods on clean images with very few distractions.

\section{Additional Ablations}

Apart from the main contribution of the paper, we found a few other training techniques beneficial to our model. First, a larger batch size helps to stablise the focal loss. We normalise the focal loss by the number of positive logits, which in itself is a very unstable statistic. Increasing the batch size from 4 to 32 results in roughly 0.8~mAP improvement. Second, using AdamW~\cite{ilya2018} instead of SGD contributes about 1~mAP to our model's performance. We attribute this improvement to the similarity between graphical models and transformers~\cite{vaswani2017}, for which AdamW is the \textit{de facto} choice of optimiser. Last, we observe a further 1~mAP improvement from fine-tuning the backbone.

\end{document}